\documentclass{bmvc2k}
\usepackage{array}
\usepackage{multirow}

\title{Mish: A Self Regularized Non-Monotonic Activation Function}

\addauthor{Diganta Misra}{mishradiganta91@gmail.com}{1}

\addinstitution{
Landskape\\
KIIT, Bhubaneswar, India
}

\runninghead{Misra}{Mish activation function}

\begin{document}

\maketitle

\begin{abstract}
 We propose \textit{Mish}, a novel self-regularized non-monotonic activation function which can be mathematically defined as: $f(x)=x\tanh(softplus(x))$. As activation functions play a crucial role in the performance and training dynamics in neural networks, we validated experimentally on several well-known benchmarks against the best combinations of architectures and activation functions. We also observe that data augmentation techniques have a favorable effect on benchmarks like ImageNet-1k and MS-COCO across multiple architectures. For example, Mish outperformed Leaky ReLU on YOLOv4 with a CSP-DarkNet-53 backbone on average precision (AP\textsubscript{50}\textsuperscript{val}) by 2.1$\%$ in MS-COCO object detection and ReLU on ResNet-50 on ImageNet-1k in Top-1 accuracy by $\approx$1$\%$ while keeping all other network parameters and hyperparameters constant. Furthermore, we explore the mathematical formulation of Mish in relation with the Swish family of functions and propose an intuitive understanding on how the first derivative behavior may be acting as a regularizer helping the optimization of deep neural networks. Code is publicly available at \url{https://github.com/digantamisra98/Mish}.
\end{abstract}

\section{Introduction}
\label{sec:intro}
Activation functions are non-linear point-wise functions responsible for introducing non-linearity to the linear transformed input in a layer of a neural network. The choice of activation function is imperative for understanding the performance of a neural network. The process of applying an activation function in a layer of a neural network can be mathematically realized as $z=g(y)=g(\sum_{i}w_{i}x_{i} + b)$ where $z$ is the output of the activation function $g(y)$. In early literature, Sigmoid and TanH activation functions were extensively used, which subsequently became ineffective in deep neural networks. A less probability inspired, unsaturated piece-wise linear activation known as Rectified Linear Unit (ReLU) \cite{nair2010rectified,krizhevsky2012imagenet} became more relevant and showed better generalization and improved speed of convergence compared to Sigmoid and TanH. 

Although ReLU demonstrates better performance and stability compared to TanH and Sigmoid, it is not without weaknesses. One of which is popularly known as Dying ReLU, which is experienced through a gradient information loss caused by collapsing the negative inputs to zero. Over the years, many activation functions have been proposed which improve performance and address the shortcomings of ReLU, which include Leaky ReLU \cite{maas2013rectifier}, ELU \cite{clevert2015fast}, and SELU \cite{klambauer2017self}. Swish \cite{ramachandran2017searching}, which can be defined as $f(x)=xsigmoid(\beta x)$, proved to be a more robust activation function showcasing strong improvements in results as compared to that of ReLU. The smooth, continuous profile of Swish proved essential in better information propagation as compared to ReLU in deep neural network architectures. 

In this work, we propose \textbf{Mish}, a novel self regularized non-monotonic activation function inspired by the self gating property of Swish. Mish is mathematically defined as: $f(x)=x\tanh(softplus(x))$. We evaluate and find that Mish tends to match or improve the performance of neural network architectures as compared to that of Swish, ReLU, and Leaky ReLU across different tasks in Computer Vision.

\begin{figure}
	\centering
	\begin{tabular}{ccc}
	\includegraphics[width=5.5cm]{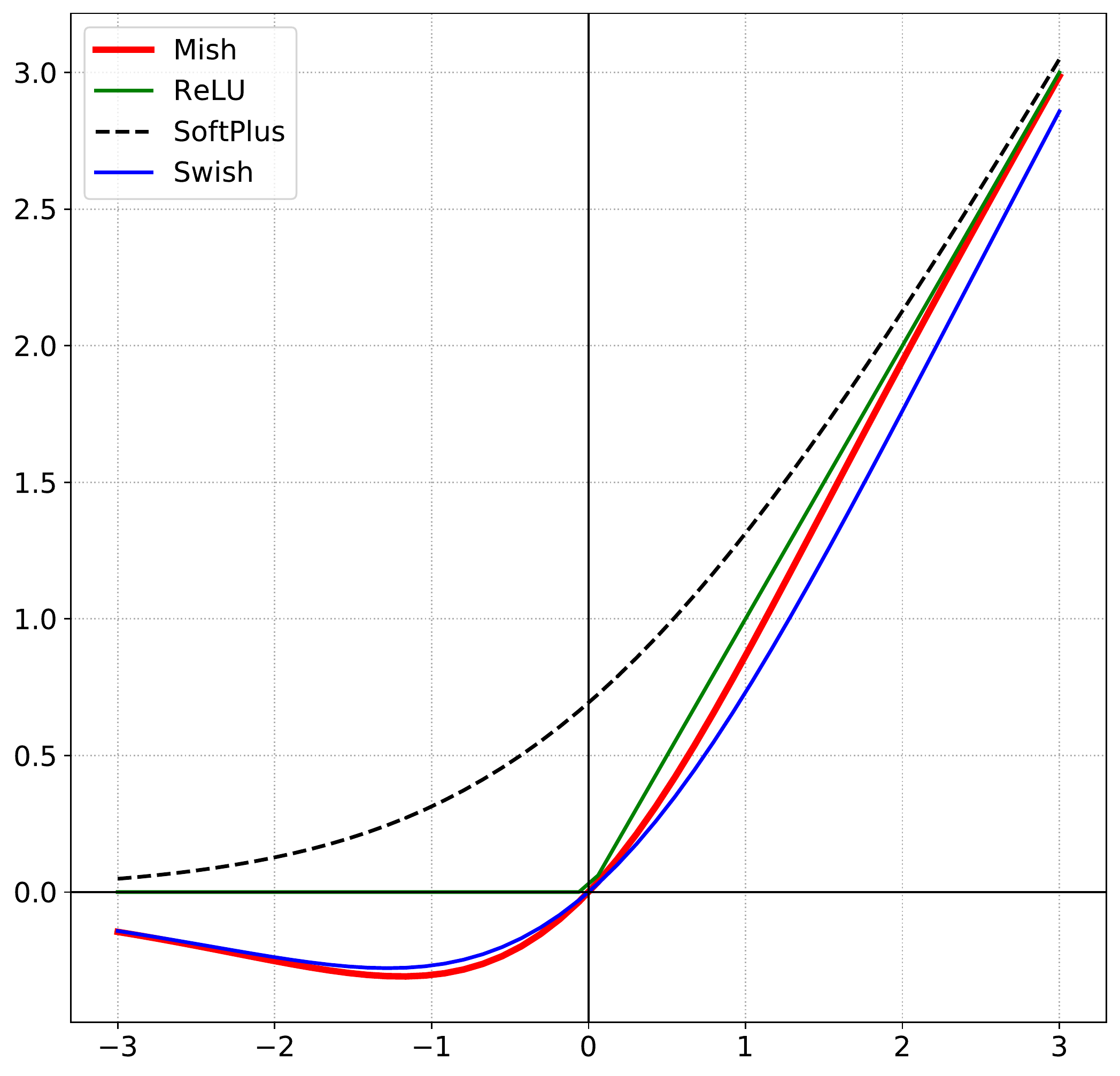}&
	\includegraphics[width=5.5cm]{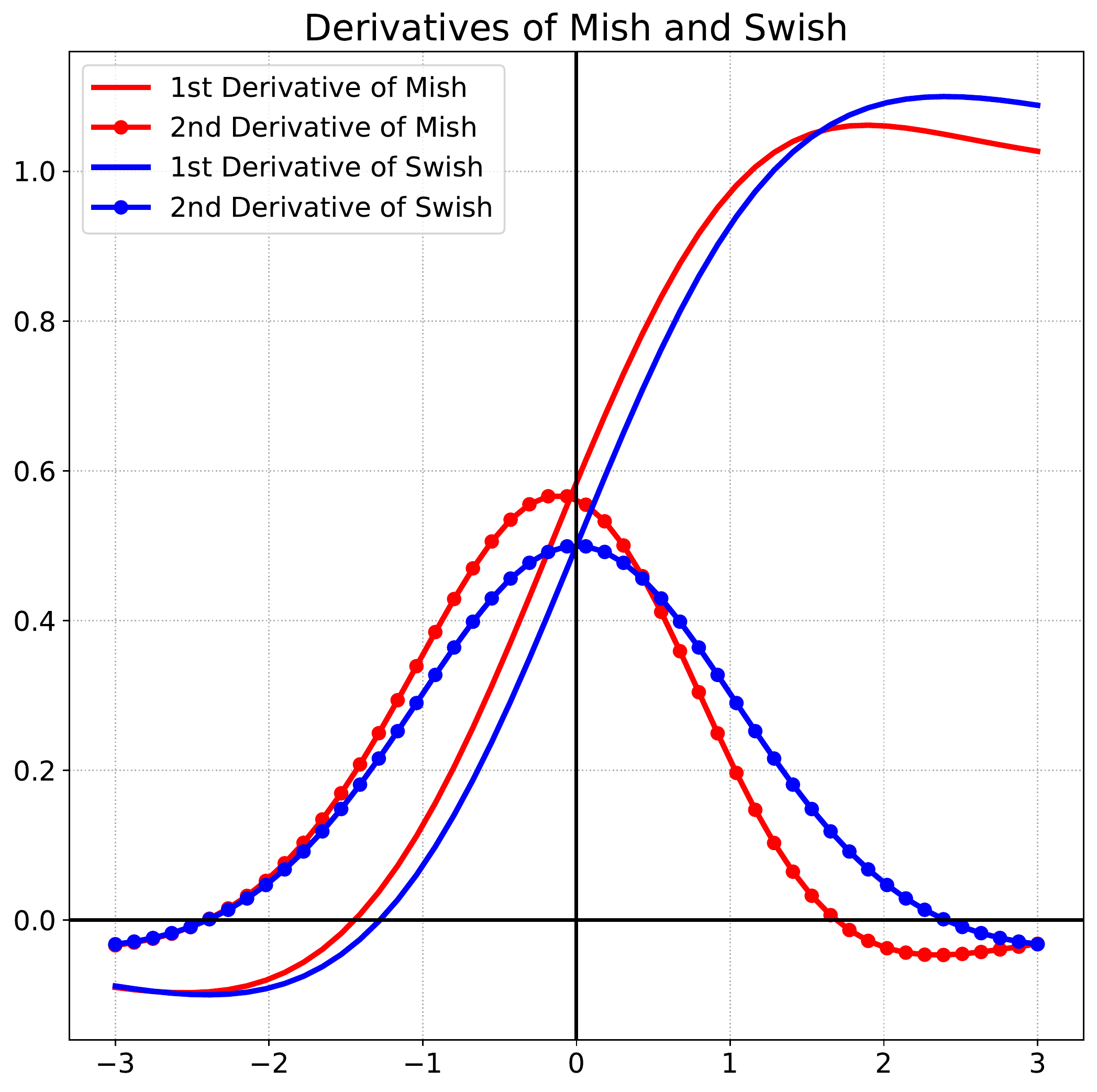}\\
	(a)&(b)
	\end{tabular}
\caption{(a) Graph of Mish, ReLU, SoftPlus, and Swish activation functions. As illustrated, Mish and Swish are closely related with both having a distinctive negative concavity unlike ReLU, which accounts for preservation of small negative weights. (b) The ${1}^{st}$ and ${2}^{nd}$ derivatives of Mish and Swish activation functions.}
\label{fig:Mish}
\end{figure}

\section{Motivation}
\label{sec:motivation}

\begin{figure}
	\centering
	\begin{tabular}{ccc}
		\includegraphics[width=5.75cm]{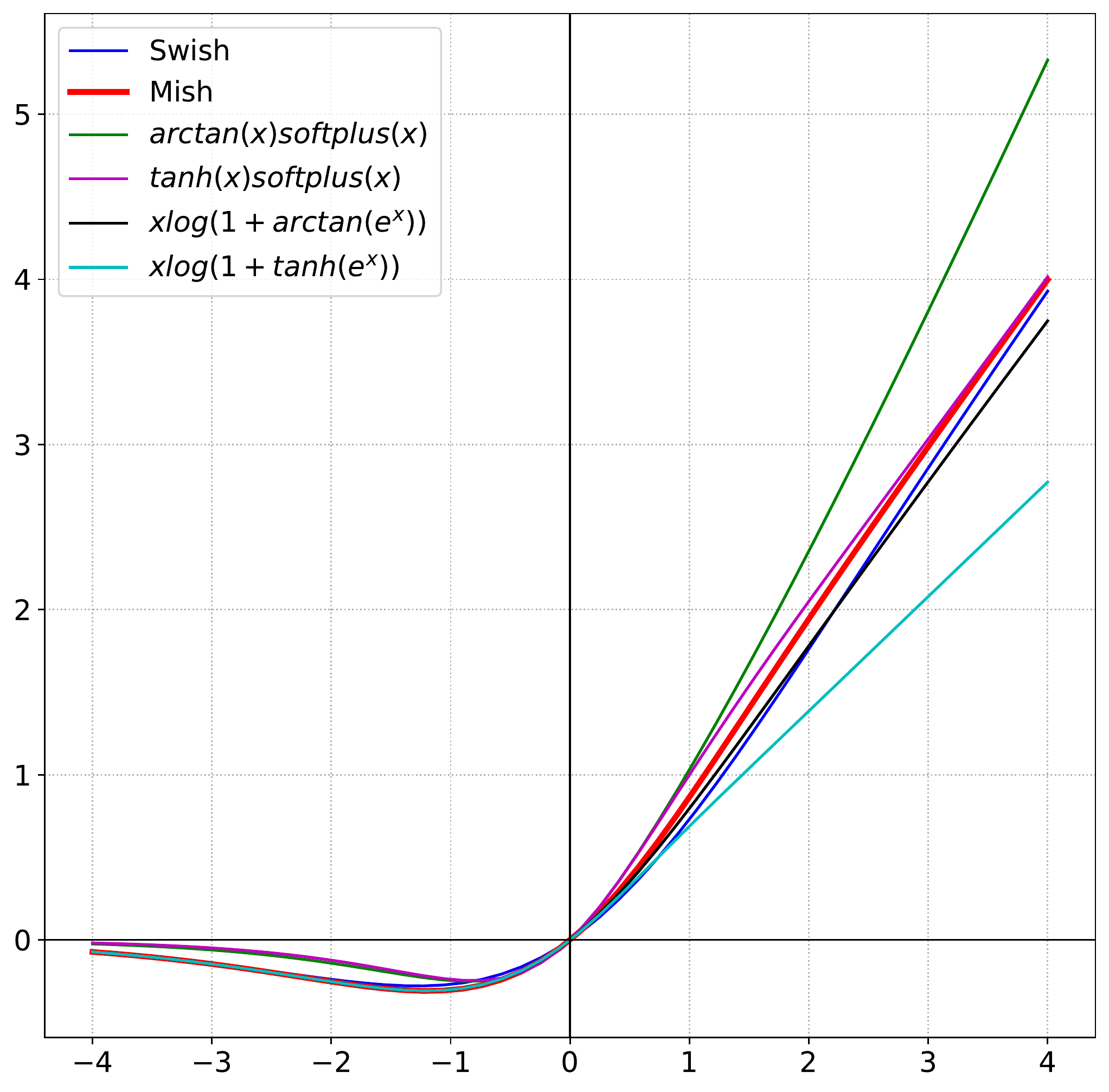}&
		\includegraphics[width=5.75cm]{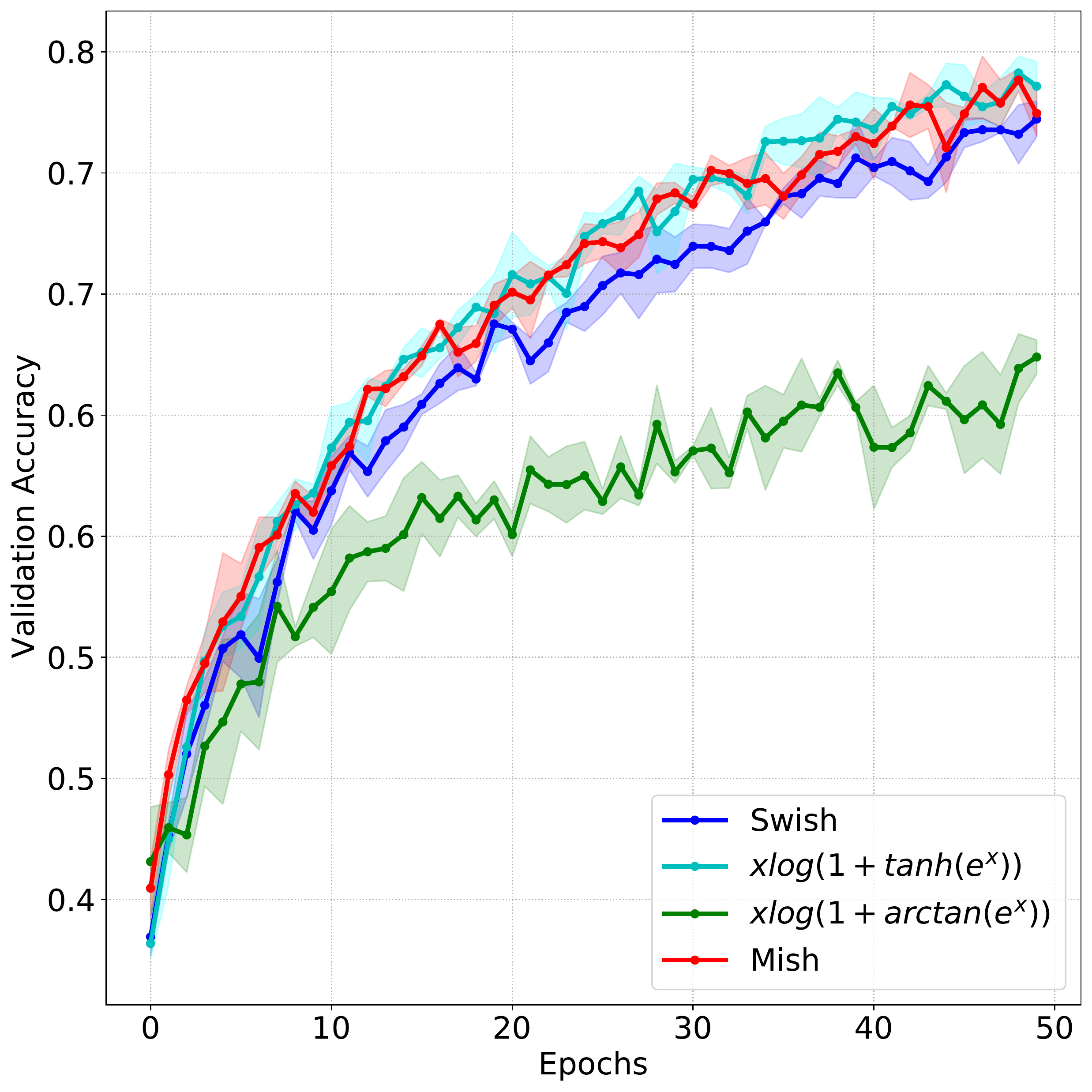}\\
		(a)&(b)
	\end{tabular}
	\caption{(a) Graph of Mish, Swish, and similar validated experimental functions. (b) Training curve of a six-layered CNN on CIFAR-10 on different validated activation functions.}
	\label{fig:candidates}
\end{figure}

Across theoretical research into activation functions, those sharing properties similar to Swish, which includes non-monotonicity, ability to preserve small negative weights, and a smooth profile, have been a recurring discussion. For instance, Gaussian Error Linear Units (GELU) \cite{hendrycks2016gaussian} is a popular activation function which has similar properties to that of Swish and is actively used in the GPT-2 architecture \cite{radford2019language} for synthetic text generation. Swish was discovered by a Neural Architecture Search (NAS) \cite{zoph2016neural} over the space of the non-linear functions by a controlled search agent. An RNN-controller was used as the agent which generated a new candidate function at each step, for a total of 10K steps, which were then evaluated on CIFAR-10 classification task using a ResNet-20 defined with that candidate function as its activation function. The design of Mish, while influenced by the work performed by Swish, was found by systematic analysis and experimentation over the characteristics that made Swish so effective. When studying similarly behaved functions like Swish, as illustrated in Fig.~\ref{fig:candidates} (a), which include $\arctan(x)softplus(x)$, $\tanh(x)softplus(x)$, $x\log(1+\arctan({e}^{x}))$ and $x\log(1+\tanh({e}^{x}))$, where $softplus(x) = \ln(1+{e}^{x})$, from our ablation study we determined Mish consistently outperforms the aforementioned functions along with Swish and ReLU.

We used a standard six-layered deep convolution neural network architecture to validate each of the experimental activation functions earlier defined on the CIFAR-10 image classification task. The networks were trained for three runs, each for 50 epochs with RMSProp as the optimizer. As shown in Fig.~\ref{fig:candidates} (b), we found that Mish performed better than the other validated functions. Although it can be observed that $x\log(1+\tanh({e}^{x}))$ performed at par to Mish, we noted that its training is often unstable and, in many cases, leads to divergence in deeper architectures. We observed similar unstable training issues for $\arctan(x)softplus(x)$ and $\tanh(x)softplus(x)$. While all of the validated functions have a similar shape, Mish proves to be consistently better in terms of performance and stability. 

While not evident at first sight, Mish is closely related to Swish, as it can be observed in the first derivative:
\begin{equation}
f'(x) = {sech}^{2}(softplus(x))xsigmoid(x) + \frac{f(x)}{x}
\end{equation}
\begin{equation}
= \Delta(x)swish(x)+\frac{f(x)}{x} 
\end{equation}
where  $softplus(x) = \ln(1+{e}^{x})$ and  $sigmoid(x) = 1/(1+{e}^{-x})$. 
\\

From experimental observations, we speculate that the $\Delta(x)$ parameter acts like a preconditioner, making the gradient smoother. Preconditioning has been extensively discussed and used in general optimization problems where the preconditioner, in case of gradient descent \cite{bottou2010large, li2017preconditioned} is the inverse of a symmetric positive definite matrix ($H^{-1}_{k}$) which is applied to modify the geometry of the objective function to increase the rate of convergence \cite{axelsson1986rate}. Intuitively, preconditioning makes the objective function much smoother and thus making it easier to optimize. The $\Delta(x)$ parameter mimics the behavior of a preconditioner. It provides a strong regularization effect and helps make gradients smoother, which corresponds to easier to optimize function contour, which is possibly why Mish outperforms Swish in increasingly deep and complex neural net architectures. 

Mish, additionally, similar to Swish, is non-monotonic, smooth, and preserves a small amount of negative weights. These properties account for the consistent performance and improvement when using Mish in-place of Swish in deep neural networks.

\section{Mish}
\label{sec:mish}

\begin{figure}
	\centering
	\includegraphics[width=0.9\linewidth]{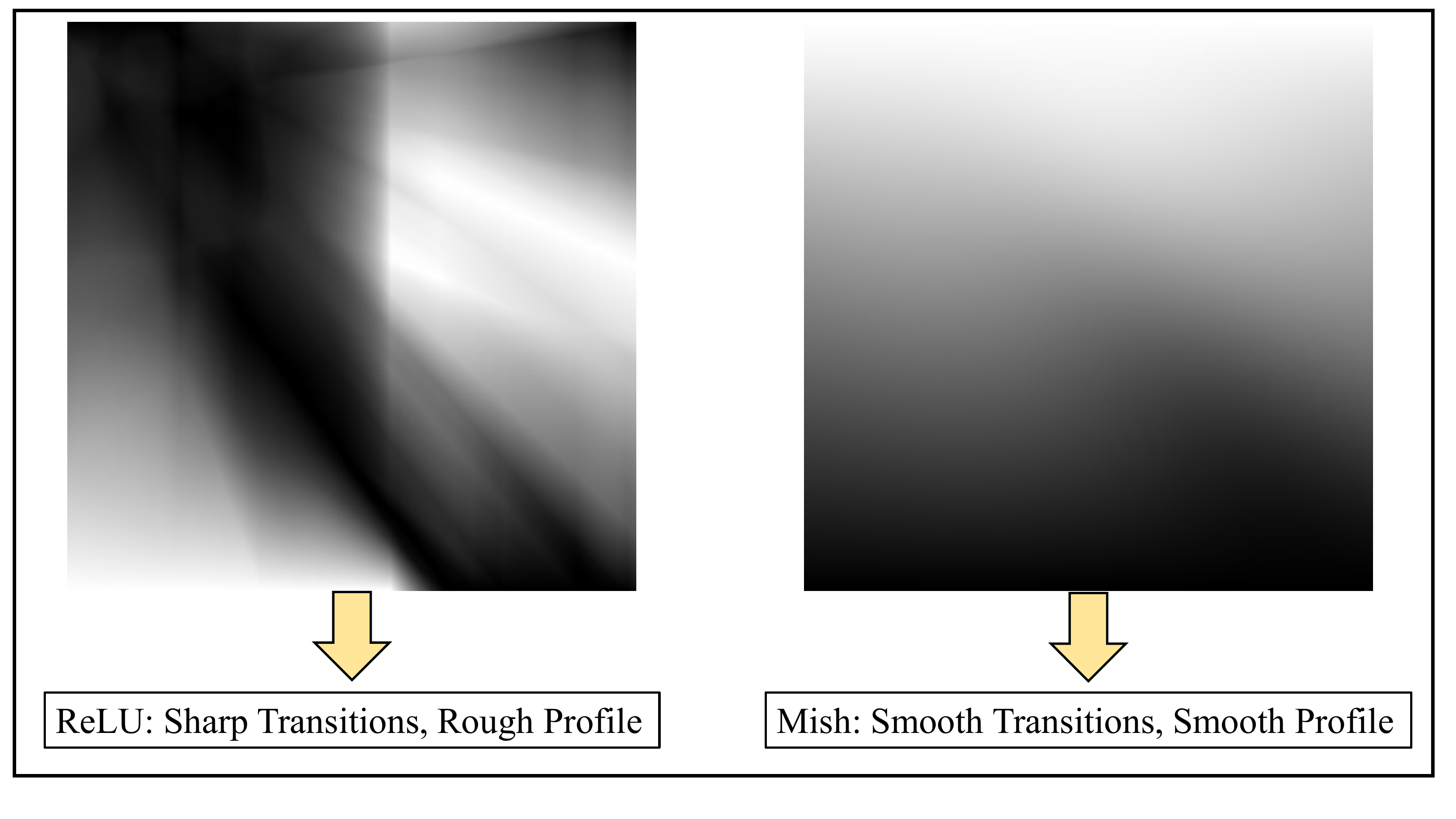}
	\caption{Comparison between the output landscapes of ReLU and Mish activation function}
	\label{fig:landscape}
\end{figure}

\begin{figure}
	\centering
	\includegraphics[width=0.9\linewidth]{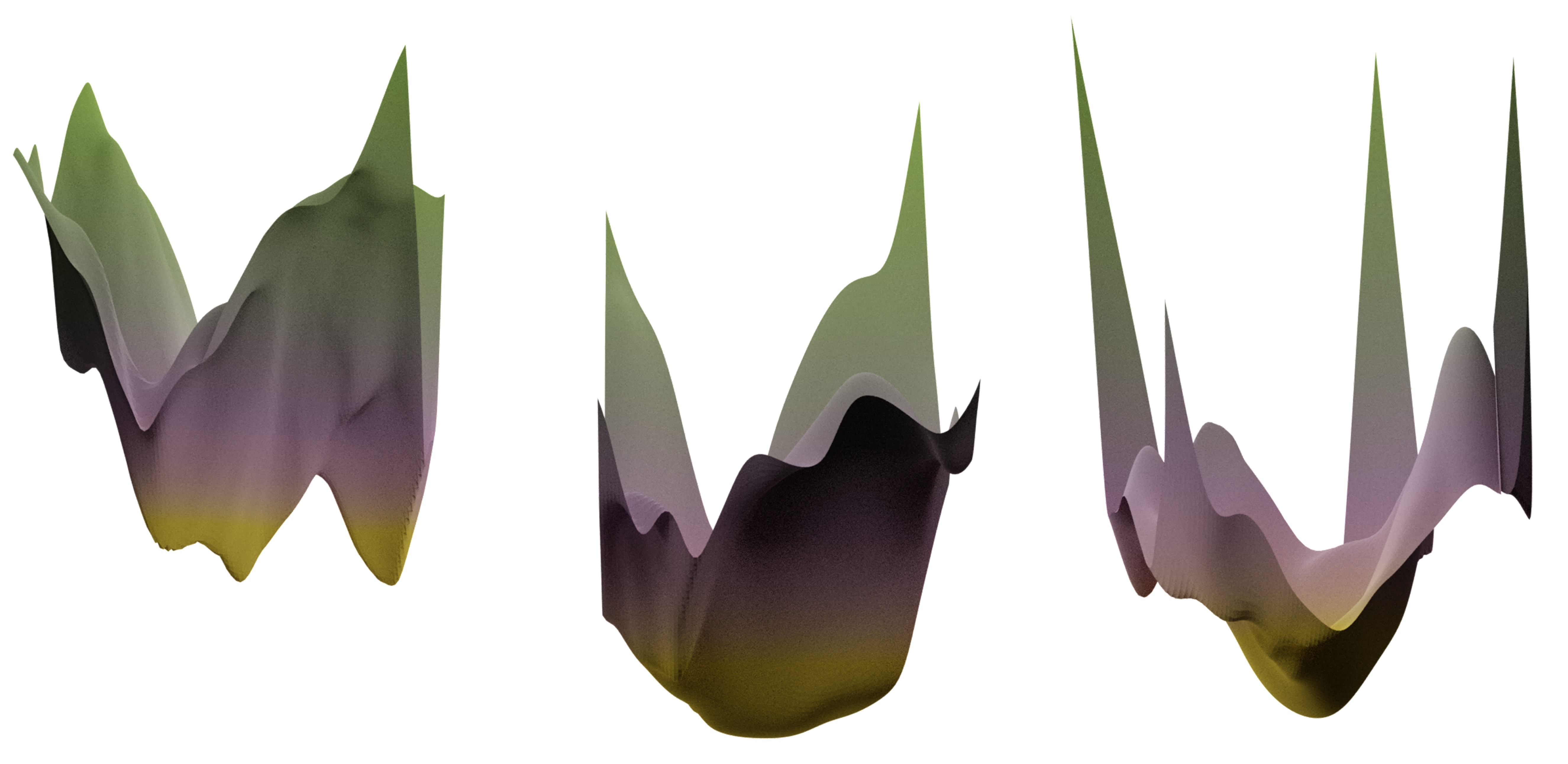}
	\caption{Comparison between the loss landscapes of (from left to right): (a) ReLU, (b) Mish and (c) Swish activation function for a ResNet-20 trained for 200 epochs on CIFAR-10.}
	\label{fig:loss}
\end{figure}

Mish, as visualized in Fig.~\ref{fig:Mish} (a), is a smooth, continuous, self regularized, non-monotonic activation function mathematically defined as:
\begin{equation}
f(x)=x\tanh(softplus(x))=x\tanh(\ln(1+e^{x}))
\end{equation}
Similar to Swish, Mish is bounded below and unbounded above with a range of [$\approx$ -0.31, $\infty$). The $1^{st}$ derivative of Mish, as shown in Fig.~\ref{fig:Mish} (b), can be defined as:
\begin{equation}
f'(x)=\frac{{e}^{x}\omega}{{\delta}^{2}}
\end{equation}
where, $\omega=4(x+1)+4{e}^{2x}+{e}^{3x}+{e}^{x}(4x+6)$ and $\delta=2{e}^{x}+{e}^{2x}+2$. Inspired by Swish, Mish uses the Self-Gating property where the non-modulated input is multiplied with the output of a non-linear function of the input. Due to the preservation of a small amount of negative information, Mish eliminated by design the preconditions necessary for the Dying ReLU phenomenon. This property helps in better expressivity and information flow. Being unbounded above, Mish avoids saturation, which generally causes training to slow down due to near-zero gradients \cite{glorot2010understanding} drastically. Being bounded below is also advantageous since it results in strong regularization effects. Unlike ReLU, Mish is continuously differentiable, a property that is preferable because it avoids singularities and, therefore, undesired side effects when performing gradient-based optimization. 

Having a smooth profile also plays a role in better gradient flow, as shown in Fig.~\ref{fig:landscape}, where the output landscapes of a five-layered randomly initialized neural network with ReLU and Mish are visualized.  The landscapes were generated by passing in the co-ordinates to a five-layered randomly initialized neural network which outputs the corresponding scalar magnitude. The output landscape of ReLU has a lot of sharp transitions as compared to the smooth profile of the output landscape of Mish. Smoother output landscapes suggest smooth loss landscapes \cite{li2018visualizing}, which help in easier optimization and better generalization, as demonstrated in Fig.~\ref{fig:loss}. 

We observed the loss landscapes \cite{li2018visualizing} of a ResNet-20 \cite{he2016deep} equipped with ReLU, Mish, and Swish activation functions with each trained for 200 epochs for the image classification task on the CIFAR-10 dataset. We used a multi-step learning rate policy with the SGD optimizer for training the networks. As shown in Fig.~\ref{fig:loss}, the loss landscape for the ResNet-20 equipped with Mish is much smoother and conditioned as compared to that of ReLU and Swish activation function. Mish has a wider minima which improves generalization compared to that of ReLU and Swish, with the former having multiple local minimas. Additionally, Mish obtained the lowest loss as compared to the networks equipped with ReLU and Swish, and thus, validated the preconditioning effect of Mish on the loss surface.

\subsection{Ablation Study on CIFAR-10 and MNIST}

\begin{figure}[h]
	\begin{tabular}{ccc}
		\includegraphics[width=4cm]{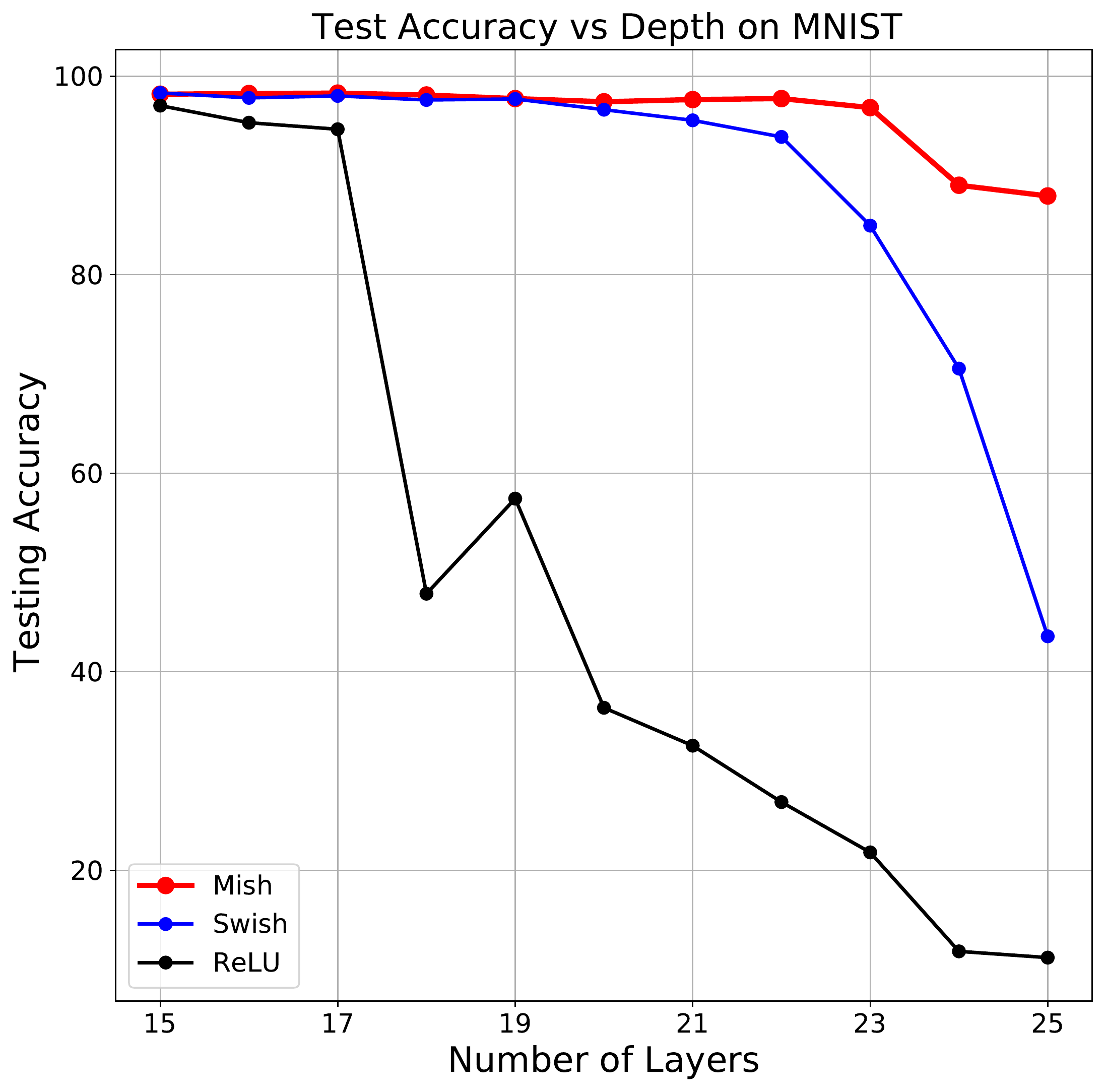}&
		\includegraphics[width=4cm]{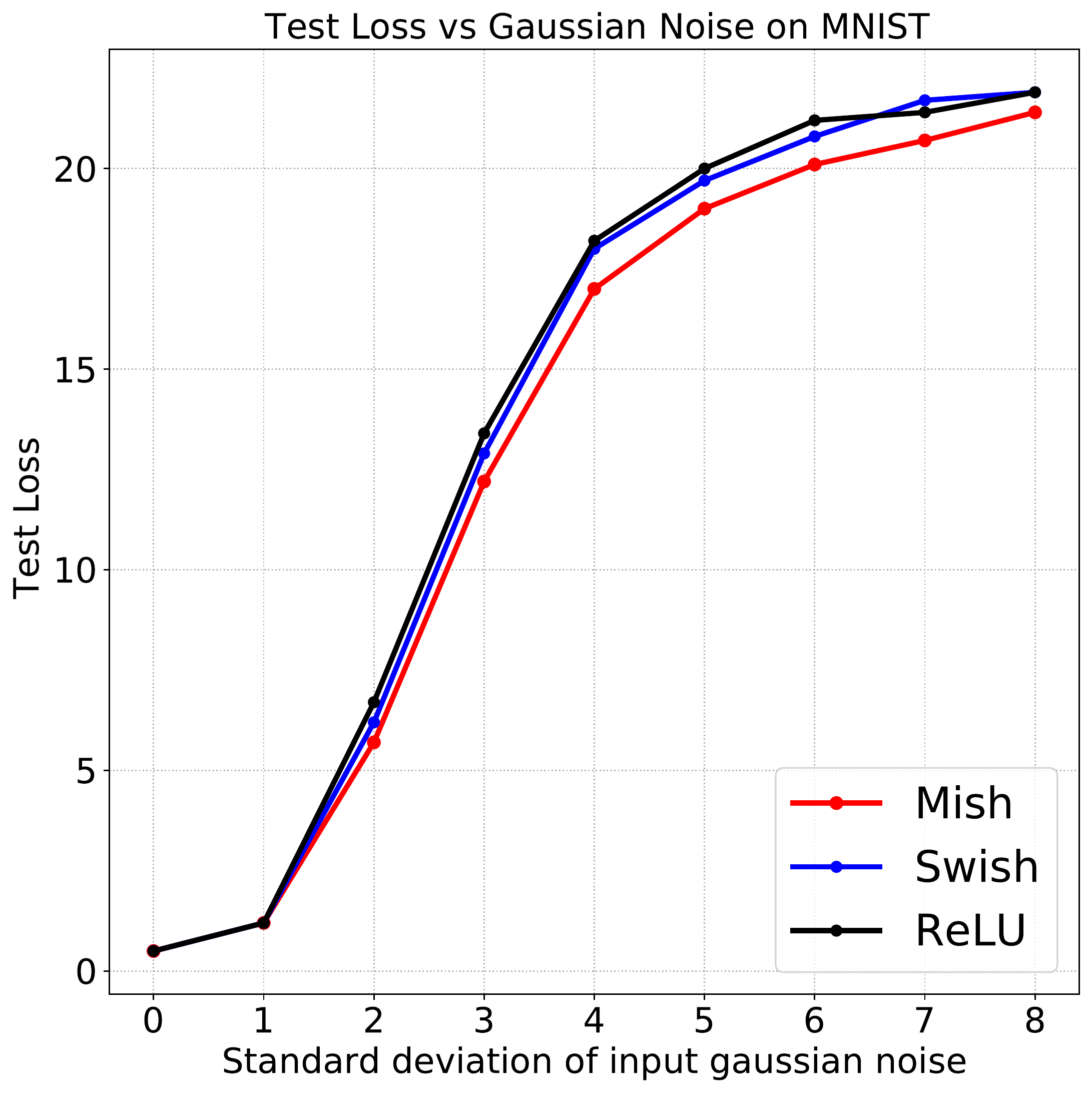}&
		\includegraphics[width=3.6cm]{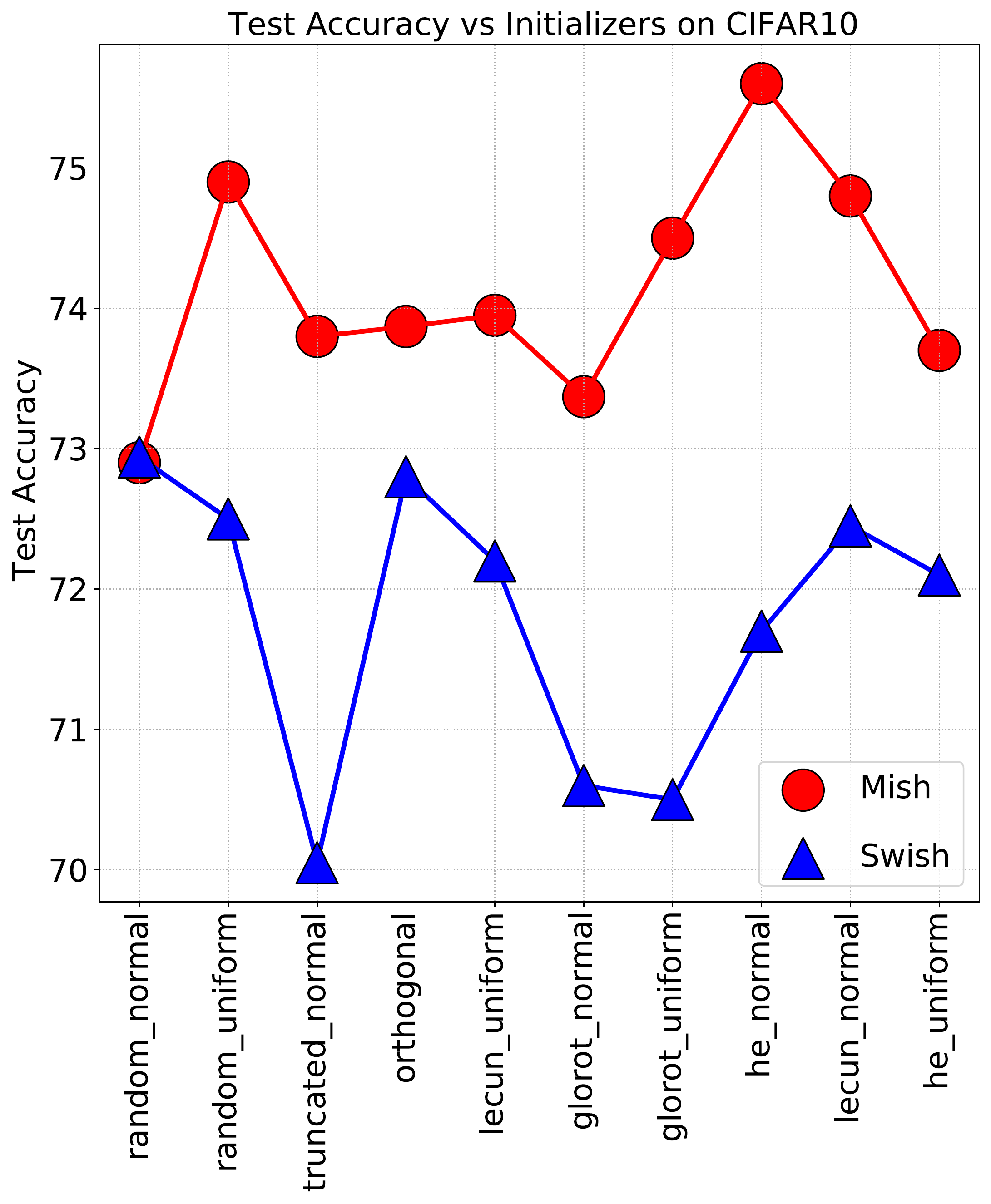}\\
		(a)&(b)&(c)
	\end{tabular}
	\caption{(a) Comparison between Mish, Swish, and ReLU activation functions in terms of test accuracy with increasing depth of the neural network on the MNIST dataset.  (b) Comparison between Mish, Swish, and ReLU activation functions in terms of test loss with increasing input gaussian noise on the MNIST dataset. (c) Comparison between Mish and Swish activation functions in terms of test accuracy with different weight initialization strategies on the CIFAR-10 dataset.}
	\label{fig:hyperparams}
\end{figure}

 Hyperparameters, including the depth of the network, type of weight initialization, batch size, learning rate, and optimizer used in the training process, have significant unique effects. We manipulate different hyper-parameters to observe their effects on the performance of ReLU, Swish, and Mish activation functions. Firstly, we observe the effect of increasing the number of layers of a neural network with ReLU, Swish, and Mish on the test accuracy. For the task, we used the MNIST dataset \cite{lecun2010mnist} and trained fully connected networks of linearly increasing depth. Each layer was initialized with 500 neurons, while Residual Units \cite{he2016deep} were not used since they allow the training of arbitrary deep networks. We used Batch Normalization \cite{ioffe2015batch} layers to decrease the dependence on initialization along with Dropout \cite{srivastava2014dropout} of 25$\%$. The network was optimized using SGD \cite{bottou2010large} with a batch size of 128. For a fair comparison, the same learning rate was maintained for the three networks with ReLU, Swish, and Mish. As shown in Fig.~\ref{fig:hyperparams} (a), post fifteen layers, there was a sharp decrease in accuracy for both Swish and ReLU, while Mish maintained a significantly higher accuracy in large models where optimization becomes difficult. This property was later validated in ImageNet-1k \cite{deng2009imagenet} experiments in Section~\ref{subsec:imagenet}, where Mish performed superior to Swish in increasingly large networks.

We also evaluated the robustness of Mish in noisy input conditions where the input MNIST data was corrupted with additive zero-centered Gaussian Noise with linearly increasing standard deviation. We used a five-layered convolution neural network architecture optimized using SGD for this task. Fig.~\ref{fig:hyperparams} (b) demonstrates the consistently better loss with varying intensity of Input Gaussian Noise with Mish as compared to ReLU and Swish.

Initializers \cite{glorot2010understanding} play a crucial role in the performance of a neural network. We observed the performance of Mish and Swish using different weight initializers, including Glorot initializer \cite{glorot2010understanding}, LeCun normal initializer \cite{lecun2012efficient}, and He uniform variance scaling initializer \cite{he2015delving}, in a six-layered convolution neural network. Fig.~\ref{fig:hyperparams} (c) demonstrates the consistent positive difference in the performance of Mish compared to Swish while using different initializers.

\section{Benchmarks}
\label{sec:benchamrks}

We evaluated Mish against more than ten standard activation functions on different models and datasets. Our results show, especially in computer vision tasks like image classification and object detection, Mish consistently matched or exceeded the best performing network. We also recorded multiple runs to observe the statistical significance of our results. Along with the vanilla settings, we also validated the performance of Mish when coupled with various state of the art data augmentation techniques like CutMix and label smoothing. 

\subsection{Statistical Analysis}

\begin{table}[h]
	\begin{center}
		\begin{tabular}{|l|c|c|c|}
			\hline
			Activation & $\mu_{acc}$ & $\mu_{loss}$ & $\sigma_{acc}$ \\
			\hline\hline
			Mish    & \textbf{87.48$\%$} & 4.13$\%$ & 0.3967 \\
			Swish \cite{ramachandran2017searching} & 87.32$\%$ & 4.22$\%$ & 0.414 \\
			GELU \cite{hendrycks2016gaussian} & 87.37$\%$ & 4.339$\%$ & 0.472 \\
			ReLU \cite{nair2010rectified,krizhevsky2012imagenet} & 86.66$\%$ & 4.398$\%$ & 0.584 \\
			ELU \cite{clevert2015fast} & 86.41$\%$ & 4.211$\%$ & 0.3371 \\
			Leaky ReLU \cite{maas2013rectifier} & 86.85$\%$ & \textbf{4.112$\%$} & 0.4569  \\
			SELU \cite{klambauer2017self} & 83.91$\%$ & 4.831$\%$ & 0.5995 \\
			SoftPlus & 83$\%$ & 5.546$\%$ & 1.4015 \\
			SReLU \cite{jin2016deep} & 85.05$\%$ & 4.541$\%$ & 0.5826 \\
			ISRU \cite{carlile2017improving} & 86.85$\%$ & 4.669$\%$ & \textbf{0.1106}\\
			TanH & 82.72$\%$ & 5.322$\%$ & 0.5826\\
			RReLU \cite{xu2015empirical} & 86.87$\%$ & 4.138$\%$ & 0.4478\\
			\hline
		\end{tabular}
	\end{center}
	\caption{Statistical results of different activation functions on image classification of CIFAR-10 dataset using a Squeeze Net for 23 runs.}
	\label{tab:p-values}
\end{table}

To evaluate the statistical significance and consistency of the performance obtained by Mish activation function compared to baseline activation functions, we calculate and compare the mean test accuracy, mean test loss, and standard deviation of test accuracy for CIFAR-10 \cite{krizhevsky2009learning} classification task using a Squeeze Net \cite{hu2018squeeze}. We experimented for 23 runs, each for 50 epochs using the Adam optimizer \cite{kingma2014adam} and changing the activation functions while keeping every other network parameter constant. Table.~\ref{tab:p-values} shows Mish outperforms other activation functions with the highest mean accuracy ($\mu_{acc}$), second-lowest mean loss ($\mu_{loss}$), and third-lowest standard deviation of accuracy ($\sigma_{acc}$).

\subsection{CIFAR-10}

We compare the performance of different baseline activation functions on the image classification task of CIFAR-10 dataset \cite{krizhevsky2009learning} using different standard neural network architectures by just swapping the activation functions and keeping every other network parameter and training parameter constant. We evaluate the performance of Mish as compared to ReLU and Swish on various standard network architectures, including Residual Networks \cite{he2016deep}, Wide Residual Networks \cite{zagoruyko2016wide}, Shuffle Net \cite{zhang2018shufflenet}, Mobile Nets \cite{howard2017mobilenets}, Inception Network \cite{szegedy2015going}, and Efficient Networks \cite{tan2019efficientnet}. Table.~\ref{tab:cifar10} shows that Mish activation function consistently outperforms ReLU and Swish activation functions across all the standard architectures used in the experiment, with often providing 1$\%$ to 3$\%$ performance improvement over the baseline ReLU enabled network architectures. 

\begin{table}[h]
	\begin{center}
		\begin{tabular}{|l|c|c|c|r|}
			\hline
			Architecture & Mish & Swish & ReLU\\
			\hline\hline
			ResNet-20 \cite{he2016deep} & \textbf{92.02$\%$} & 91.61$\%$ & 91.71$\%$ \\
			WRN-10-2 \cite{zagoruyko2016wide} & \textbf{86.83$\%$} & 86.56$\%$ & 84.56$\%$ \\
			SimpleNet \cite{hasanpour2016lets} & \textbf{91.70$\%$} & 91.44$\%$ & 91.16$\%$ \\
			Xception Net \cite{chollet2017xception} & \textbf{88.73$\%$} & 88.56$\%$ & 88.38$\%$ \\
			Capsule Net \cite{sabour2017dynamic} & \textbf{83.15$\%$} & 82.48$\%$ & 82.19$\%$ \\
			Inception ResNet v2 \cite{szegedy2015going} & \textbf{85.21$\%$} & 84.96$\%$ & 82.22$\%$ \\
			DenseNet-121 \cite{huang2017densely} & \textbf{91.27$\%$} & 90.92$\%$ & 91.09$\%$ \\
			MobileNet-v2 \cite{howard2017mobilenets} & \textbf{86.25$\%$} & 86.08$\%$ & 86.05$\%$ \\
			ShuffleNet-v1 \cite{zhang2018shufflenet} & \textbf{87.31$\%$} & 86.95$\%$ & 87.04$\%$ \\
			Inception v3 \cite{szegedy2015going} & \textbf{91.19$\%$} & 91.17$\%$ & 90.84$\%$ \\
			Efficient Net B0 \cite{tan2019efficientnet} & \textbf{80.73$\%$} & 79.37$\%$ & 79.31$\%$ \\
			\hline
		\end{tabular}
	\end{center}
	\caption{Comparison between Mish, Swish, and ReLU activation functions based on test accuracy on image classification of CIFAR-10 across various network architectures.}
	\label{tab:cifar10}
\end{table}

\subsection{ImageNet-1k}
\label{subsec:imagenet}

Additionally, we compare Mish with Leaky ReLU \cite{maas2013rectifier} and Swish for ImageNet 2012 dataset classification task. ImageNet \cite{deng2009imagenet} is considered to be one of the most challenging and significant classification tasks in the domain of computer vision. ImageNet comprises of 1.28 million training images distributed across 1,000 classes. We use the validation set comprising of 50,000 images to evaluate the performance of the trained networks. We trained the networks using the DarkNet framework \cite{darknet13} on an AWS EC2 p3.16xlarge instance comprising of 8 Tesla V100 GPUs for a total number of 8 million training steps with batch size, mini-batch size, initial learning rate, momentum, and weight decay set at 128, 32, 0.01, 0.9, and 5e-4 respectively.

\begin{table}[h]
	\begin{center}
		\resizebox{\textwidth}{!}{\begin{tabular}{|c|c|cc|cc|cc|cc|cc|}
				\hline
				Model & Data Augmentation & \multicolumn{2}{c|}{LReLU/ ReLU \textsuperscript{$\dagger$}} & \multicolumn{2}{c|}{Swish} & \multicolumn{2}{c|}{Mish} \\
				& & Top-1 & Top- 5 & Top-1 & Top- 5 & Top-1 & Top- 5 \\
				\hline\hline
				ResNet-18 \cite{he2016deep} & No & 69.8$\%$\textsuperscript{$\dagger$} & 89.1$\%$\textsuperscript{$\dagger$} & \textbf{71.2$\%$} & \textbf{90.1$\%$} & \textbf{71.2$\%$} & 89.9$\%$ \\
				\hline
				ResNet-50 \cite{he2016deep} & No & 75.2$\%$\textsuperscript{$\dagger$} & 92.6$\%$\textsuperscript{$\dagger$} & 75.9$\%$ & 92.8$\%$ & \textbf{76.1$\%$} & \textbf{92.8$\%$} \\
				\hline
				SpineNet-49 \cite{du2019spinenet} & Yes & 77.0$\%$\textsuperscript{$\dagger$} & 93.3$\%$\textsuperscript{$\dagger$} & 78.1$\%$ & 94$\%$ & \textbf{78.3$\%$} & \textbf{94.6$\%$} \\
				\hline
				PeleeNet \cite{wang2018pelee} & No & 70.7$\%$ & 90.0$\%$ & \textbf{71.5$\%$} & \textbf{90.7$\%$} & 71.4$\%$ & 90.4$\%$ \\
				\hline
				CSP-ResNet-50 \cite{wang2019cspnet} & Yes & 77.1$\%$ & 94.1$\%$ & - & - & \textbf{78.1$\%$} & \textbf{94.2$\%$} \\
				\hline
				CSP-DarkNet-53 \cite{bochkovskiy2020yolov4} & Yes & 77.8$\%$ & 94.4$\%$ & - & - & \textbf{78.7$\%$} & \textbf{94.8$\%$} \\
				\hline
				CSP-ResNext-50 \cite{wang2019cspnet} & No & 77.9$\%$ & 94.0$\%$ & 64.5$\%$ & 86$\%$ & \textbf{78.9$\%$} & \textbf{94.5$\%$} \\
				CSP-ResNext-50 \cite{wang2019cspnet} & Yes & 78.5$\%$ & 94.8$\%$ & - & - & \textbf{79.8$\%$} & \textbf{95.2$\%$} \\
				\hline
		\end{tabular}}
	\end{center}
	\caption{Comparison between Mish, Swish, ReLU and Leaky ReLU activation functions on image classification of ImageNet-1k dataset across various standard architectures. Data Augmentation indicates the use of CutMix, Mosaic, and Label Smoothing. \textsuperscript{$\dagger$} indicate scores for ReLU.}
	\label{tab:imagenet}
\end{table}

In Table.~\ref{tab:imagenet}, we compare the Top-1 and Top-5 accuracy of Mish against ReLU, Leaky ReLU, and Swish on PeleeNet \cite{wang2018pelee} , Cross Stage Partial ResNet-50 \cite{wang2019cspnet}, and ResNet-18/50 \cite{he2016deep}. Mish consistently outperforms the default Leaky ReLU/ ReLU on all the four network architectures with a 1$\%$ increase in Top-1 Accuracy over Leaky ReLU in CSP-ResNet-50 architecture. Although Swish provides marginally stronger result in PeleeNet as compared to Mish, we investigate further of the inconsistency of the performance of Swish in a larger model where we compare Swish, Mish and ReLU in a CSP-ResNext-50 model \cite{wang2019cspnet, xie2017aggregated} where Swish decreases the Top-1 accuracy by 13.4$\%$ as compared to Leaky ReLU while Mish improves the accuracy by 1$\%$. This shows that Swish cannot be used in every architecture and has drawbacks in especially large complex models like ResNext based models. We also combine different data augmentation techniques like CutMix \cite{yun2019cutmix} and Label Smoothing (LS) \cite{muller2019does} to improve the baseline scores of CSP-ResNet-50, CSP-DarkNet-53 \cite{bochkovskiy2020yolov4} and CSP-ResNext-50 models. The results suggest that Mish is more consistent and generally guarantees performance increase in almost any neural network for ImageNet classification.

\subsection{MS-COCO Object Detection}

Object detection \cite{girshick2014rich} is a fundamental branch of computer vision that can be categorized as one of the tasks under visual scene understanding. In this section, we present our experimental results on the challenging Common Objects in Context (MS-COCO) dataset \cite{lin2014microsoft}. We report the mean average precision (mAP-50/ mAP@0.5) on the COCO test-dev split, as demonstrated in Table. ~\ref{tab:coco}. We report our results for two models, namely, CSP-DarkNet-53 \cite{bochkovskiy2020yolov4} and CSP-DarkNet-53+PANet+SPP \cite{bochkovskiy2020yolov4, he2015spatial}, where we retrained the backbone network from scratch by replacing the activation function from ReLU to Mish. 

We also validate our results by using various data augmentation strategies, including CutMix \cite{yun2019cutmix} , Mosaic \cite{bochkovskiy2020yolov4}, self adversarial training (SAT) \cite{bochkovskiy2020yolov4}, Dropblock regularization \cite{ghiasi2018dropblock} and Label Smoothing \cite{muller2019does} along with Mish. As per the results demonstrated in Table. ~\ref{tab:coco}, simply replacing ReLU with Mish in the backbone improved the mAP@0.5 for CSP-DarkNet-53 and CSP-DarkNet-53+PANet+SPP by $0.4\%$. For CSP-DarkNet-53, we achieve state of the art mAP@0.5 of 65.7$\%$ at a real-time speed of 65 FPS on Tesla V100. Additionally, CSP-DarkNet-53 was used as the backbone with a Yolov3 detector \cite{redmon2018yolov3} as its object detection head. We use multi-input weighted residual connections (MiWRC) \cite{tan2019efficientdet} in the backbone and train the model with a cosine annealing scheduler \cite{loshchilov2016sgdr}. We also eliminate grid sensitivity and use multiple anchors for single ground truth for the detector. Experiments were done on a single GPU to enable multi-scale training with default parameters, including epochs, initial learning rate, weight decay, and momentum set at 500500, 0.01, 5e-4, and 0.9, respectively. 

\begin{table}[h]
	\begin{center}
		\resizebox{\textwidth}{!}{\begin{tabular}{|c|c|c|c|c|}
				\hline
				Model & Size & Data Augmentation & ReLU & Mish \\
				\hline\hline
				CSP-DarkNet-53 \cite{bochkovskiy2020yolov4} & (512 x 512) & No & 64.5$\%$ & \textbf{64.9$\%$} \\
				CSP-DarkNet-53 \cite{bochkovskiy2020yolov4} & (608 x 608) & No & - & \textbf{65.7$\%$} \\
				\hline
				CSP-DarkNet53+PANet+SPP \cite{bochkovskiy2020yolov4, he2015spatial} & (512 x 512) & Yes & 64.5$\%$ & \textbf{64.9$\%$}\\
				\hline
		\end{tabular}}
	\end{center}
	\caption{Comparison between ReLU and Mish activation functions on object detection on MS-COCO dataset.}
	\label{tab:coco}
\end{table}

We provide further comparative results using the YOLOv4 \cite{bochkovskiy2020yolov4} detector, as demonstrated in Table. ~\ref{tab:coco1}. Using Mish, we observed a consistent 0.9$\%$ to 2.1$\%$ improvement in the AP\textsubscript{50}\textsuperscript{val} on test size of 736. We evaluated three variants of YOLOv4, which are: YOLOv4\textsubscript{pacsp}, YOLOv4\textsubscript{pacsp-s}, and YOLOv4\textsubscript{pacsp-x}. All three variants use a CSP-DarkNet-53 \cite{wang2019cspnet} and CSP-PANet in the backbone coupled with a CSP-SPP \cite{he2015spatial} (Spatial Pyramid Pool) module where the latter two variants denote the tiny and extra-large variant of YOLOv4\textsubscript{pacsp}.

\begin{table}[h]
	\begin{center}
		\resizebox{\textwidth}{!}{\begin{tabular}{|c|c|c|c|c|c|c|c|}
				\hline
				Detector & Activation & AP\textsuperscript{val} & AP\textsubscript{50}\textsuperscript{val} & AP\textsubscript{75}\textsuperscript{val} & AP\textsubscript{S}\textsuperscript{val} & AP\textsubscript{M}\textsuperscript{val} & AP\textsubscript{L}\textsuperscript{val} \\
				\hline\hline
				\multirow{2}{*}{YOLOv4\textsubscript{pacsp-s}} & Leaky ReLU & 36.0$\%$ & 54.2$\%$ & 39.4$\%$ &	18.7$\%$ &	41.2$\%$	& 48.0$\%$ \\
				& Mish & \textbf{37.4$\%$} &	\textbf{56.3$\%$} &	\textbf{40.0$\%$} &	\textbf{20.9$\%$} &	\textbf{43.0$\%$} &	\textbf{49.3$\%$}\\
				\hline
				\multirow{2}{*}{YOLOv4\textsubscript{pacsp}} & Leaky ReLU & 46.4$\%$ &	64.8$\%$ &	\textbf{51.0$\%$} &	28.5$\%$ &	51.9$\%$ &	\textbf{59.5$\%$}	 \\
				& Mish & \textbf{46.5$\%$} &	\textbf{65.7$\%$} &	50.2$\%$ &	\textbf{30.0$\%$} &	\textbf{52.0$\%$} &	59.4$\%$\\
				\hline
				\multirow{2}{*}{YOLOv4\textsubscript{pacsp-x}} & Leaky ReLU & 47.6$\%$ &	66.1$\%$ &	52.2$\%$ &	29.9$\%$ &	53.3$\%$ &	61.5$\%$	 \\
				& Mish & \textbf{48.5$\%$} &	\textbf{67.4$\%$} &	\textbf{52.7$\%$} &	\textbf{30.9$\%$} &	\textbf{54.0$\%$} &	\textbf{62.0$\%$}\\
				\hline
		\end{tabular}}
	\end{center}
	\caption{Comparison between Leaky ReLU and Mish activation functions on object detection on MS-COCO 2017 dataset with a test image size of 736 x 736.}
	\label{tab:coco1}
\end{table}

\subsection{Stability, Accuracy, and Efficiency Trade-off}

Mish is a novel combination of three activation functions, which are TanH, SoftPlus, and the identity function. In practical implementation, a threshold of 20 is enforced on Softplus, which makes the training more stable and prevents gradient overflow. Due to the increased complexity, there is a trade-off between the increase in accuracy while using Mish and the increase in computational cost. We address this concern by optimizing Mish using a CUDA based implementation, which we call Mish-CUDA which is based on PyTorch \cite{paszke2019pytorch}.

\begin{table}[h]
	\begin{center}
		\begin{tabular}{|c|c|c|c|c|c|}
			\hline
			Activation & Data Type & Forward Pass & Backward Pass \\
			\hline\hline
			ReLU & fp16 & 223.7$\mu s$ $\pm$ 1.026$\mu s$ & 312.1$\mu s$ $\pm$ 2.308$\mu s$ \\
			SoftPlus & fp16 & 342.2$\mu s$ $\pm$ 38.08$\mu s$ & 488.5$\mu s$ $\pm$ 53.75$\mu s$ \\
			Mish & fp16 & 658.8$\mu s$ $\pm$ 1.467$\mu s$ & 1.135$ms$ $\pm$ 4.785$\mu s$ \\
			Mish-CUDA & fp16 & 267.3$\mu s$ $\pm$ 1.852$\mu s$ & 345.6$\mu s$ $\pm$ 1.875$\mu s$ \\
			\hline
			ReLU & fp32 & 234.2$\mu s$ $\pm$ 621.8$ns$  & 419.3$\mu s$ $\pm$ 1.238$\mu s$ \\
			SoftPlus & fp32 & 255.1$\mu s$ $\pm$ 753.6$ns$ & 420.2$\mu s$ $\pm$ 631.4$ns$ \\
			Mish & fp32 & 797.4$\mu s$ $\pm$ 1.094$\mu s$ & 1.689$ms$ $\pm$ 1.222$\mu s$ \\
			Mish-CUDA & fp32 & 282.9$\mu s$ $\pm$ 876.1$ns$ & 496.3$\mu s$ $\pm$ 1.781$\mu s$ \\
			\hline
		\end{tabular}
	\end{center}
	\caption{Comparison between the runtime for the forward and backward passes for ReLU, SoftPlus, Mish and Mish-CUDA activation functions for floating point-16 and floating point-32 data.}
	\label{tab:speed}
\end{table}

In Table.~\ref{tab:speed}, we show the speed profile comparison between the forward pass (FWD) and backward pass (BWD) on floating-point 16 (FP16) and floating-point 32 (FP32) data for ReLU, SoftPlus, Mish, and Mish-CUDA. All runs were performed on an NVIDIA GeForce RTX-2070 GPU using standard benchmarking practices over 100 runs, including warm-up and removing outliers.

Table.~\ref{tab:speed} shows the significant reduction in computational overhead of Mish by using the optimized version Mish-CUDA which shows no stability issues, mirrors the learning performance of the original baseline Mish implementation and is even faster than native PyTorch Softplus implementation in single precision, making it more feasible to use Mish in deep neural networks. Mish can be further optimized using the exponential equivalent of the TanH term to accelerate the backward pass, which involves the derivative computation.

\section{Conclusion}

In this work, we propose a novel activation function, which we call Mish. Even though Mish shares many properties with Swish and GELU like unbounded positive domain, bounded negative domain, non-monotonic shape, and smooth derivative, Mish still provides under most experimental conditions, better empirical results than Swish, ReLU, and Leaky ReLU. We expect that a hyperparameter search with Mish as a target may improve upon our results. We also observed that the state of the art data augmentation techniques like CutMix and other proven ones like Label Smoothing behave consistently with the expectations. 

Future work includes optimizing Mish-CUDA to reduce the computational overhead further, evaluating the performance of the Mish activation function in other state of the art models on various tasks in the domain of computer vision, and obtaining a normalizing constant as a parameter for Mish which can reduce the dependency on using Batch Normalization layers. We believe it is of theoretical importance to investigate the contribution of the $\Delta(x)$ parameter at the first derivative and understand the underlying mechanism on how it may be acting as a regularizer. A clear understanding of the behavior and conditions governing this regularizing term could motivate a more principled approach to constructing better performing activation functions. 

\section{Acknowledgements}

The author would like to dedicate this work to the memory of his late grandfather, Prof. Dr. Fakir Mohan Misra. The author would also like to offer sincere gratitude to everyone who supported during the timeline of this project including Sparsha Mishra, Alexandra Deis from X – The Moonshot Factory, Ajay Uppili Arasanipalai from University of Illinois - Urbana Champaign (UIUC), Himanshu Arora from Montreal Institute for Learning Algorithms (MILA), Javier Ideami, Federico Andres Lois from Epsilon, Alexey Bochkovskiy, Chien-Yao Wang, Thomas Brandon, Soumik Rakshit from DeepWrex, Less Wright, Manjunath Bhat from Indian Institute of Technology - Kharagpur (IIT-KGP), Miklos Toth and many more including the Fast.ai team, Weights and Biases community and everyone at Landskape.

\bibliography{egbib}
\end{document}